\documentclass{article}
\usepackage{cite}
\usepackage{amsmath,amssymb,amsfonts}
\usepackage{algorithmic}
\usepackage{graphicx}
\usepackage{bm}
\usepackage{textcomp}
\usepackage{xcolor}
\usepackage{multirow}
\usepackage{tablefootnote}
\usepackage{algorithmic,algorithm,xpatch}

\def\BibTeX{{\rm B\kern-.05em{\sc i\kern-.025em b}\kern-.08em
    T\kern-.1667em\lower.7ex\hbox{E}\kern-.125emX}}
\usepackage{ifthen}
\newboolean{showcomments}
\setboolean{showcomments}{true}
\newcommand{\indra}[1]{\ifthenelse{\boolean{showcomments}}
{ \textcolor{red}{(Indra says:  #1)}}{}} 

\newcommand{\sai}[1]{\ifthenelse{\boolean{showcomments}}
{ \textcolor{red}{(Sai says:  #1)}}{}}

\begin{document}

\title{Virtual Battery Parameter Identification using Transfer Learning based Stacked Autoencoder
\thanks{The authors would like to thank U.S. Department of Energy for supporting this work under the ENERGISE program (contract DE-AC02-76RL01830).\newline This work has been accepted in \textit{17th IEEE International Conference on Machine Learning and Applications}, Orlando, 2018. Copyright 2018 by the author(s).}
}

\author{
 Indrasis~Chakraborty, Sai~Pushpak~Nandanoori and Soumya~Kundu\\
Optimization and Control Group, Pacific Northwest National Laboratory \\ Richland, USA\\
{\tt\small \{indrasis.chakraborty, saipushpak.n, soumya.kundu\}@pnnl.gov}
}
\date{}
\maketitle

\begin{abstract}
Recent studies have shown that the aggregated dynamic flexibility of an ensemble of thermostatic loads can be modeled in the form of a virtual battery. The existing methods for computing the virtual battery parameters require the knowledge of the first-principle models and parameter values of the loads in the ensemble. In real-world applications, however, it is likely that the only available information are end-use measurements such as power consumption, room temperature, device on/off status, etc., while very little about the individual load models and parameters are known. We propose a transfer learning based deep network framework for calculating virtual battery state of a given ensemble of flexible thermostatic loads, from the available end-use measurements. This proposed framework extracts first order virtual battery model parameters for the given ensemble. We illustrate the effectiveness of this novel framework on different ensembles of ACs and WHs. 
\end{abstract}

\section{Introduction}

In recent years, deep learning has been gaining popularity among researchers due to its inherent nature of imitating human brain learning process. Specifically, several research works show the applicability of deep learning in capturing representative information from raw dataset using multiple nonlinear transformations as shown in \cite{hinton2006reducing}. Deep learning based methods can model high level abstractions in data utilizing multiple processing layers, compared to shallow learning methods. In \cite{bengio2007greedy}, deep learning methods are used for simplifying a learning task from input examples. Based on scientific knowledge in the area of biology, cognitive humanoid autonomous methods with deep-learning-architecture
have been proposed and applied over the years
\cite{hinton2006fast,lecun2015deep,zhang2013denoising,bengio2015editorial,varol2015efficient}. Deep learning replaces handcrafted feature extraction by learning unsupervised features as shown in \cite{song2013hierarchical}. 
%
%
%
%
However, although there exists great potential, 
application of these deep network based frameworks are relatively sparse in power system applications. 

The last few years have seen a significant increase in integration of renewables into the electricity grid, and the intermittent nature of renewables causes uncertainty in power generation. Moreover, there is increased visibility into the operations of thermostatically controllable loads (TCLs) such as air conditioners (ACs) and electric water heaters (WHs) due to advancements in power electronics, communication capabilities that enable remote monitoring/controlling of TCLs. With increased renewable penetration, these advancements allow TCLs to provide grid ancillary services in the form of demand response, such as frequency regulation \cite{Molina_Garcia:2011, callaway2011achieving, bashash2011modeling, koch2011modeling, hao2015aggregate, zhang2013aggregated, kundu2011modeling, pushpak2018prioritized}. The thermostatic loads such as ACs and electric WHs can be characterized as `energy-driven' loads, because end-use quality of service is reliant on the energy consumption over a duration. Availability of these loads to address demand response is strongly influenced by their dynamics. While modeling and coordinating each load individually in a power systems network is intractable for a grid operator, a reduced-order model to characterize the dynamic flexibility of an ensemble of loads is more amenable to be integrated into grid operation and planning tools. To this effect, recent studies have proposed the \textit{`virtual battery'} (or, `generalized battery') models for an ensemble of flexible thermostatic loads.  

Most recent works on virtual battery (VB) model identification can be seen in \cite{ mathieu2015arbitraging, hao2015aggregate, hughes2016identification, hao2017optimal}. These works include characterizing a VB model for a wide range of systems from small residential TCLs to complex systems such as commercial building heating, ventilating, and air-conditioning loads. Similar to a real battery, the VB also has self-dissipation, energy capacity, and power (charge/discharge) limits as parameters. Existing state of the art methods on identification of VB parameters for an ensemble of thermostatic loads can be either via closed-form (albeit, ad-hoc) expressions or via optimization-based techniques. Both of these require the knowledge of the individual load models and parameters. In real-world applications, however, it is expected that very little about the individual load models and parameters are known. The only available information are the end-use measurements such as power consumption, room temperature, device on/off status, etc. In this work, we propose an alternative, data-driven, deep neural network framework of characterizing the aggregate flexibility of ensemble of ACs and WHs using the available end-use measurements. Since the training of the deep networks is an offline process requiring high computational efforts, it is not desirable to retrain the network if number of TCLs in the ensemble change over time due to changing availability of end-use appliances. Henceforth, we propose a transfer learning based approach to identify the VB parameters of the new ensemble with minimal retraining.

The main contribution of this paper is to develop a novel transfer learning based stacked autoencoder framework for representing virtual battery state of a given ensemble of TCLs. We also propose a novel one-dimensional convolution based LSTM network for calculating rest of the VB parameters, utilizing the previously calculated VB state from the stacked autoencoder. Finally, the VB parameters are numerically identified for two different ensembles of ACs and WHs. 

The organization of this paper is as follows. In Section \ref{sec2} we introduce stacked autoencoder along with the respective optimization problem for its training. In Section \ref{sec3}, we introduce the training objective of a LSTM network, and we describe the proposed one-dimensional convolution operation in Section \ref{sec4}. Transfer learning for the stacked autoencoder is introduced in Section \ref{sec5}. In Section \ref{sec6}, we introduce the dynamics and the power limit calculations for any given ensemble of ACs and WHs. In Section \ref{sec7}-\ref{sec8}, we introduce a novel training process for training the convolution based LSTM network along with a detailed description of the dataset and whole deep network based framework. Numerical results are discussed in Section \ref{sec9}.

\section{Description of Stacked Autoencoder}
\label{sec2}
Autoencoder \cite{baldi2012autoencoders} (AE) is a type of deep neural network which gets trained by restricting the output values to be equal to the input values. This also indicates both input and output spaces have same dimensionality. The reconstruction error between the input and the output of network is used to adjust the weights of each layers. Therefore, the features learned by AE can well represent the input data space. Moreover, the training of AE is unsupervised, since it does not require label information.
\subsection{General setup for AE}
We have considered a supervised learning problem with a training set of $n$ (input,output) pairs $S_{n}=\{(x^{(1)},y^{(1)}),\dots,(x^{(n)},y^{(n)})\}$, that is sampled from an unknown distribution $q(X,Y)$. $X$ is a $d$ dimensional vector in $\mathbb{R}^{d}$. $Z\in\mathbb{R}^{d^{\prime}}$ is a lower ($d^{\prime}<d$) dimensional representation of $X$. $Z$ is linked to $X$ by a deterministic mapping $f_{\theta}$, where $\theta$ is a vector of trainable parameters. Now, we briefly mention the terminology associated with AE.

\textbf{Encoder:} Encoder involves a deterministic mapping $f_{\theta}$ which transforms an input vector $\bm{x}$ into hidden representation $\bm{z}$.\footnote{Throughout this paper, bold faced symbols denote vectors} $f_{\theta}$ is an affine nonlinear mapping defined as
\begin{equation}
\label{encode1}
f_{\theta}(\bm{x})=s(\bm{W}\bm{x}+\bm{b}),
\end{equation}
where $\theta=\{\bm{W},\bm{b}\}$ is a set of parameters, and $\bm{W}$ is $d^{\prime}\times d$ weight matrix, $\bm{b}$ is a bias vector of dimension $d^{\prime}$, and $s(.)$ is an activation function.

\textbf{Decoder:} The hidden dimensional representation $\bm{z}$ is mapped to dimension $d$, using mapping $g_{\theta^{\prime}}$ and represented as $\bm{\hat{y}}$. The mapping $g_{\theta^{\prime}}$ is called the decoder. Similar to $f_{\theta}$, $g_{\theta^{\prime}}$ is also an affine nonlinear mapping defined as
\begin{equation}
\label{decode1}
g_{\theta^{\prime}}(\bm{z})=s(\bm{W^{\prime}}\bm{z}+\bm{b^{\prime}}),
\end{equation}
where $\theta^{\prime}=\{\bm{W}^{\prime},\bm{b}^{\prime}\}$. Also, $\bm{\hat{y}}$ is not an exact reconstruction of $\bm{x}$, but rather in probabilistic terms as the parameters of a distribution $p(X|\hat{Y}=\bm{\hat{y}})$ that may generate $\bm{x}$ with high probability. Now, we can equate the encoded and decoded outputs as $p(X|Z=\bm{z})=p(X|\hat{Y}=g_{\theta^{\prime}}(\bm{\hat{y}}))$. Now, the reconstruction error to be optimized is $L(\bm{x},\bm{\hat{y}})\propto -\log p(\bm{x}|\bm{\hat{y}})$. For real valued $\bm{x}$, $L(\bm{x},\bm{\hat{y}})=L_{2}(\bm{x},\bm{\hat{y}})$, where $L_{2}(.,.)$ represents euclidean distance between two variables. In other words, we will use the squared error objective for training our autoencoder. For this current work, we will use affine and linear encoder along with affine and linear decoder with squared error loss. 

Furthermore, autoencoder training consists of minimizing the reconstruction error, by carrying out the optimization 
\begin{equation}
J_{AE}=\arg \min_{\theta,\theta^{\prime}}\mathbb{E}_{q^{0}(X)}[L(X,\hat{Y}(X)]\nonumber,
\end{equation}
where $\mathbb{E}_{(.)}[\times]$ denotes the expectation, and $q^{0}$ is the empirical distribution defined by samples in $S_{n}$. Now for the loss function defined before, the optimization problem can be rewritten as
\begin{equation}
J_{AE}=\arg \min_{\theta,\theta^{\prime}}\mathbb{E}_{q^{0}(X)}[\log p(X|\hat{Y})]=g_{\theta^{\prime}}(f_{\theta}(X))]\nonumber.
\end{equation}
Intuitively, objective of training an autoencoder is to minimize the reconstruction error amounts by maximizing lower bound on the shared information between $X$ and hidden representation $Z$. 
\begin{figure}[!ht]
\centering
\includegraphics[width=\columnwidth]{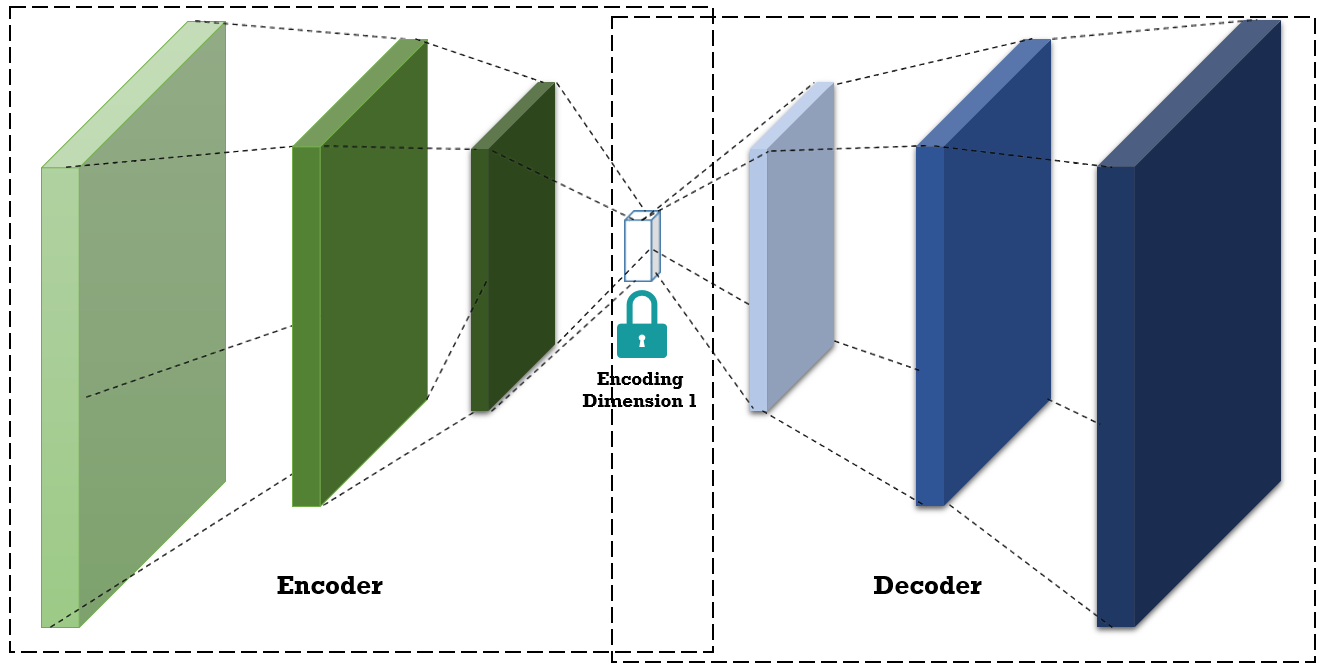} 
\caption{Schematic of the proposed SAE}
\label{fig:SAE}
\end{figure}
\subsection{Stacked Autoencoder (SAE)}
We utilize the proposed AE and stack them to initialize a deep network as the similar way of deep belief networks \cite{hinton2006reducing} or ordinary AEs (\cite{bengio2007greedy},\cite{ranzato2007automatic}, \cite{larochelle2009exploring}). Once the AEs have been properly stacked, the inner most encoding layer output as shown in schematic, Figure \ref{fig:SAE}, is considered as a virtual battery representation of the ensemble of TCLs. Furthermore, the number of layers of the stacked AEs are designed based on the reconstruction error for AE. Keeping in mind, a sudden change in dimension in both encoding and decoding layers can cause a difficulty in minimizing the reconstruction error in $J_{AE}$. The parameters of all layers is fine-tuned using a stochastic gradient descent approach \cite{bottou2010large}.

\section{Description of Long-Short-Term-Memory network}
\label{sec3}
\begin{figure}
\centering
\includegraphics[scale=0.7]
{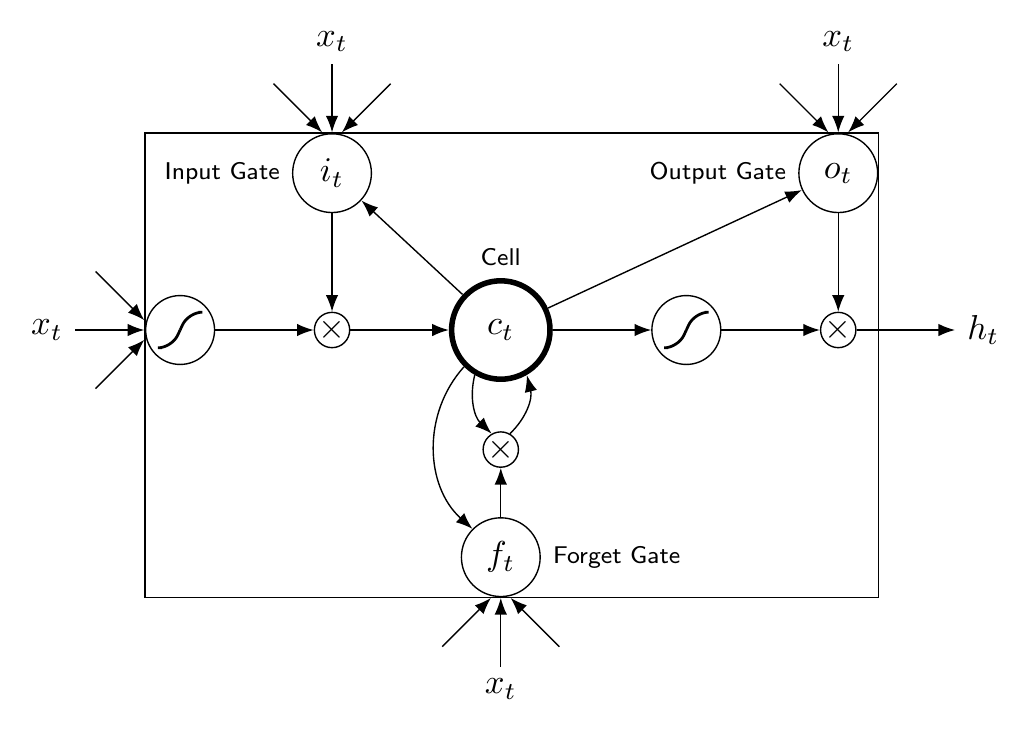} 
\caption{Schematic of one \textbf{LSTM} unit}
\label{fig:LSTM}
\end{figure}

We have used \textit{long short-term memory} \textbf{LSTM} \cite{hochreiter1997long} to learn the long range temporal dependencies of virtual battery state of a given ensemble (encoded representation of the TCL states). For a \textbf{LSTM} cell with N memory units (see Figure \ref{fig:LSTM}), at each time step, the evolution of its parameters is determined by
\begin{align*}
i_{t} & =\sigma(W_{x_{i}}x_{t}+W_{h_{i}}h_{t-1}+W_{c_{i}}c_{t-1}+b_{i_{1}}),\\
f_{t} & =\sigma(W_{x_{f}}u_{t}+W_{h_{f}}h_{t-1}+W_{c_{f}}c_{t-1}+b_{i_{f}}),\\
z_{t} & =\tanh(W_{x_{c}}x_{t}+W_{h_{c}}h_{t-1}+b_{c}),\\
c_{t} & =f_{t}\bigodot c_{t-1}+i_{t}\bigodot z_{t},\\
o_{t} & =\sigma(W_{u_{o}}u_{t}+W_{h_{o}}h_{t-1}+W_{c_{o}}c_{t-1}+b_{i_{o}}),\\
h_{t} & =o_{t}\bigodot\tanh(c_{t}),
\end{align*}
where the $W_{x_{()}}$ and $W_{h_{()}}$ terms are the respective
rectangular input and square recurrent weight matrices, $W_{c_{()}}$
are peephole weight vectors from the cell to each of the gates, $\sigma$
denotes sigmoid activation functions (applied element-wise) and the
$i_{t}$, $f_{t}$, and $o_{t}$ equations denote the input, forget,
and output gates, respectively; $z_{t}$ is the input to the cell $c_{t}$.
The output of a \textbf{LSTM} cell is $o_{t}$ and denotes pointwise vector products. The forget gate facilitates resetting the state of the \textbf{LSTM}, while the peephole connections from the cell to the gates enable accurate learning of timings. 

The goal of a \textbf{LSTM} cell training is to estimate the conditional probability $p(o_{t}|i_{t})$ where $i_{t}$ consists of a concatenated set of variables (virtual battery state of previous time steps and control input) and $o_{t}$ consists of virtual battery state of current time step. The proposed \textbf{LSTM} calculates this conditional probability by first obtaining fixed dimensional representation $v_{t}$ of the input $i_{t}$ given by the hidden state $h_{t}$. Subsequently, the conditional probability of $o_{t}$ is calculated by using the hidden state representation $h_{t}$. Now given a training dataset with input $S$ and output $T$, training of the proposed \textbf{LSTM} is done by maximizing the log probability of the training objective
\begin{equation}
\label{LSTM_obj}
J_{LSTM}=\frac{1}{|\mathcal{S}|}\sum_{(T,S)\in\mathcal{S}} \log p(T|S),
\end{equation}
where $\mathcal{S}$ denotes training set. After successful training, the forecasting is done by translating the trained \textbf{LSTM} as
\begin{equation}
\label{LSTM_obj1}
\hat{T}=\arg \max_{T} p(T|S),
\end{equation}
where $\hat{T}$ is the \textbf{LSTM} based prediction of output dataset $T$.

\begin{figure*}
\centering
\includegraphics[width=\columnwidth]
{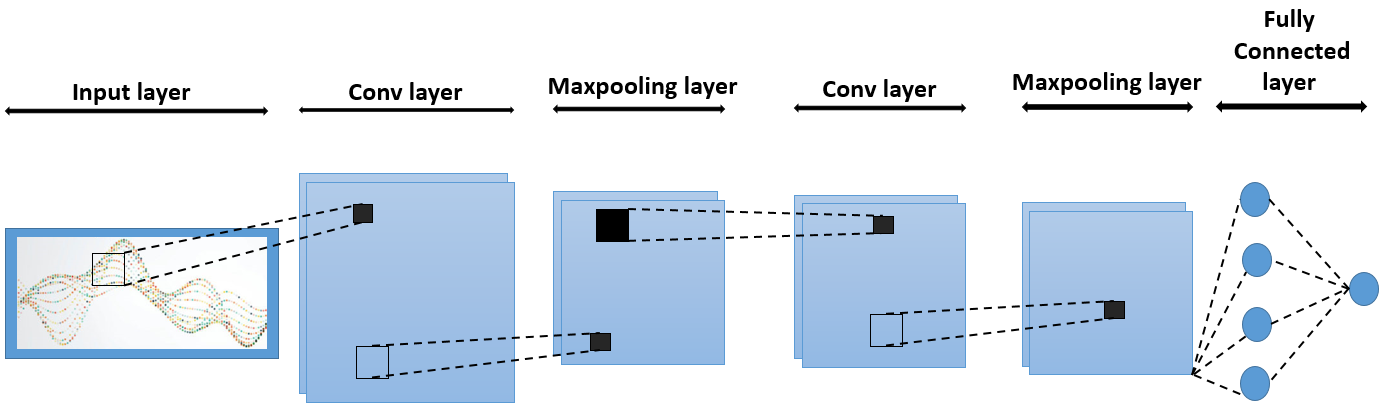} 
\caption{Schematic of proposed convolution network}
\label{fig:CNN}
\end{figure*}
\section{Description of ConvNet}
\label{sec4}
A simple convolution neural network (\textbf{ConvNet}) is a sequence of layers, and every layer of a \textbf{ConvNet} transforms one volume of activations to another through a differentiable function. In this work, we used two type of layers to build \textbf{ConvNet} architectures: a Convolution Layer and a Pooling Layer. We stack these two layers alternately to form a \textbf{ConvNet}. We can write the output of a one-dimensional \textbf{ConvNet} as follows:
\begin{itemize}
\item  One-dimensional convolution layer
\vspace{-5pt}
\begin{itemize}
\item Accepts a volume of size $W_{1}\times H_{1}\times D_{1}$, where $W_{1}$ is the batch size of the training data set
\item Requires four hyperparameters: number of filters $K_{C}$, their spatial extent $F_{C}$, length of stride $S_{C}$, and the amount of zero padding $P_{C}$
\item Produces a volume of size $W_{2}\times H_{2}\times D_{2}$, where
\vspace{-8pt}
\begin{eqnarray}
W_{2}&=&W_{1}\nonumber\\ 
H_{2}&=&(H_{1}-F_{C}+2P_{C})/S_{C}+1\nonumber \\
D_{2}&=&K_{C} \label{def_conv}
\end{eqnarray}
\item Each filter in a convolution layer introduces $F\times F\times D_{1}$ weights, and in total $(F\times F\times D_{1})\times K$ weights and $K$ biases
\end{itemize}
\item Pooling layer
\begin{itemize}
\item Accepts a volume of size $W_{2}\times H_{2}\times D_{2}$
\item Requires two hyperparameters: spatial extent $F_{P}$, and stride $S_{P}$
\item Produces a volume of size $W_{3}\times H_{3}\times D_{3}$, where
\vspace{-8pt}
\begin{eqnarray}
W_{3}&=&W_{2}\nonumber\\
H_{3}&=&(H_{2}-F_{P})/S_{P}+1\nonumber\\
D_{3}&=&D_{2}\label{def_max}
\end{eqnarray}
\item Introduces zero weights and biases, since the pooling layer computes a fixed function of the input.
\end{itemize}
\end{itemize}
Next, for our proposed \textbf{ConvNet}, we will outline the architecture as applied to predict VB state and simultaneously learn and estimate VB parameters.
\begin{itemize}
\item \textbf{Input} $[b_{s}\times l_{b}\times 1]$, where $b_{s}$ denotes the batch size of our training process (two hours, with $1$ second resolution).
\item The convolutional (\textbf{CONV}) layer computes the output of neurons that are connected to local regions in the input, each computing a dot product between their weights and a small region they are connected in the input volume. We have multiple \textbf{CONV} layers in our proposed \textbf{ConvNet}, each with a different filter size $K_{C}$. For an input layer of size $[b_{s}\times l_{b}\times 1]$, the output of \textbf{CONV} layer will be $[b_{s}\times l_{b}\times K_{C}]$.
\item The rectified linear unit (\textbf{RELU}) layer will apply an element-wise activation function, such as \textbf{$\max (0,x)$ } thresholding at zero. This leaves the size of the output volume unchanged to $[b_{s}\times l_{b}\times K_{C}]$.
\item The pooling (\textbf{POOL}) layer will perform a down-sampling operation along the spatial dimension (width, height), resulting in an output volume such as $[b_{s}\times \frac{(l_{b}-F_{P})}{S_{P}}+1 \times K_{C}]$, with a filter size of $F_{P}\times S_{P}$.
\end{itemize}

\section{Transfer learning via Net2Net for SAE}
\label{sec5}
The structure (input node numbers) for the proposed SAE depends on the number of TCLs in the ensemble. This requires retraining of the SAE if we change the number of TCL in the ensemble. We are proposing to use the developed \textbf{Net2Net} strategy \cite{chen2015net2net} where there is a change in number or type of TCLs in the ensemble. In order to explain this idea in the context of VB state modeling, we define "source system" ($S$) as an ensemble of $N$ devices (where $N$ is a defined integer)\footnote{For clarity we discuss \textbf{Net2Net} in the context of homogeneous device ensemble.}. We will further define "target system" ($T$) as an ensemble of $M$ devices (where $M\neq N$). The goal is that $S$ will provide a good internal representation of the given task for $T$ to copy and begin refining. This idea was initially presented as \textbf{FitNets} \cite{romero2014fitnets} and subsequently modified by Chen et al. in \cite{chen2015net2net} as \textbf{Net2Net}. 

We are proposing to combine two \textbf{Net2Net} strategies, namely \textbf{Net2WiderNet} and \textbf{Net2DeeperNet}\footnote{\textbf{Net2WiderNet} and \textbf{Net2DeeperNet} were first introduced by Chen et al. in \cite{chen2015net2net}}. Both of them based on initializing the "target" network to represent the same function as the "source" network. As an example, let the SAE representing $S$ is represented by a function $\bm{\hat{y}}=f(\bm{x},\theta)$, where $\bm{x}$ is input to the SAE, $\bm{\hat{y}}$ is output from the SAE (which for SAE is the reconstruction of input $x$), and $\theta$ is the trainable parameters of the SAE. We propose to choose a new set of parameters $\theta^{\prime}$ for the SAE representing $T$ such that
\begin{equation}
\forall \bm{x},f(\bm{x},\theta)=g(\bm{x},\theta^{\prime})\nonumber
\end{equation}
\subsection{Net2DeeperNet}
As the name suggests, \textbf{Net2DeeperNet} allows to transform a network into a deeper one. Mathematically, \textbf{Net2DeeperNet} replaces one layer with two layers, i.e., $h^{(i)}=\phi(h^{(i-1)T} W_{1}^{(i)})$ gets replaced by $h^{(i)}=\phi(W_{2}^{(i)T} \phi(W_{1}^{(i)T} h^{(i-1)}))$. The new weight matrix $W_{2}$ is initialized as identity matrix and get updated in the training process. Moreover, we need to ensure that $\phi(I\phi(v))=\phi(v)$ for all $v$, in order to ensure \textbf{Net2DeeperNet} can successfully replace the original network with deeper ones. 
\subsection{Net2WiderNet}
\textbf{Net2WiderNet} allows a layer to be replaced with a wider layer, meaning a layer that has more neurons (can be also narrower if needed). Suppose that layer $i$ and layer $i+1$ are both fully connected layers, and layer $i$ uses an elementwise non-linearity. To widen layer $i$, we replace $W^{(i)}$ with $W^{(i+1)}$. If layer $i$ has $m$ inputs and $n$ outputs, and layer $i+1$ has $p$ outputs, then $W^{(i)}\in\mathbb{R}^{m\times n}$ and $W^{(i+1)}\in\mathbb{R}^{n\times p}$. \textbf{Net2WiderNet}  allows to replace layer $i$ with a layer that has $q$ outputs, with $q>n$. We will introduce a random mapping function $g:\{1,2,\dots,q\}\rightarrow\{1,2,\dots,n\}$, that satisfies
\begin{eqnarray}
g(j)&=&j,j\leq n\nonumber\\
g(j)&=&\text{random sample from } {1,2,\dots,n},j>n\nonumber
\end{eqnarray}
The new weights in the network for target $T$ is given by
\begin{eqnarray}
U_{k,j}^{(i)}&=&W^{(i)}_{k,g(j)},\nonumber\\
U_{j,h}^{(i+1)}&=&\frac{1}{|\{x|g(x)=g(j)\}|} W^{(i+1)}_{g(j),h}.\nonumber
\end{eqnarray}
Here, the first $n$ columns of $W^{(i)}$ are copied directly into $U^{(i)}$. Columns $n+1$ through $q$ of $U^{(i)}$ are created by choosing a random sample as defined in $g$. The random selection is performed with replacement, so each column of $W^{(i)}$ is copied potentially many times. For weights in $U^{(i+1)}$, we must account for the replication by dividing the weight by replication factor given by $\frac{1}{|\{x|g(x)=g(j)\}|}$, so all the units have exactly the same value as the unit in the network in source $S$.
\begin{figure}
\centering
\includegraphics[width=\columnwidth]
{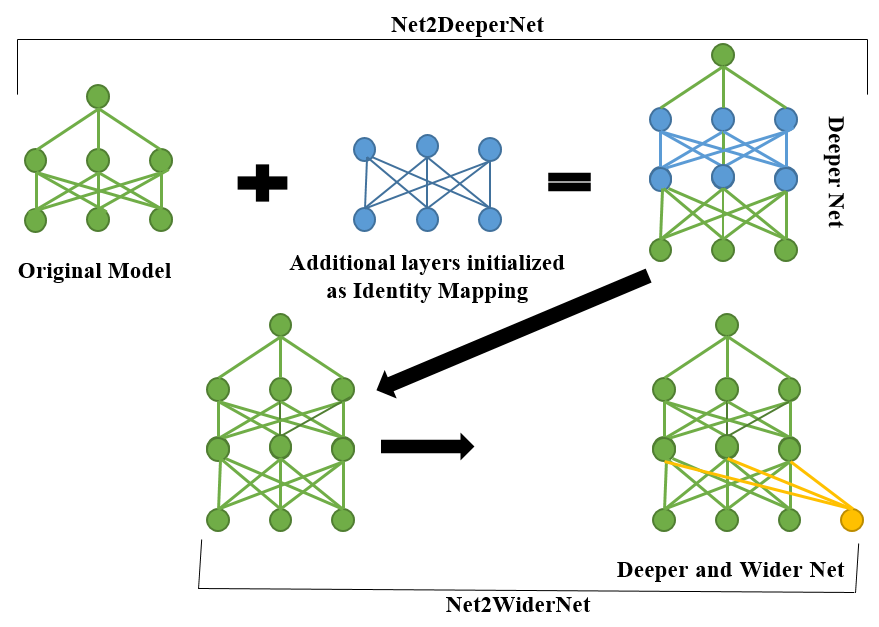} 
\caption{Proposed consecutive \textbf{Net2DeeperNet} and \textbf{Net2WiderNet} operations on original trained model}
\label{fig:Net2Net}
\end{figure}

Next, we discuss the application of this transfer learning to identify Virtual Battery (VB) parameters corresponding to an ensemble of ACs or WHs. We begin with the description of the VB model. 

\section{Virtual Battery Model}
\label{sec6}
We use the following first order system model to describe the dynamics of a VB system, 
\begin{subequations}
\begin{align}
\dot{x}(t) &= - a x(t) - u(t)\,,\quad x(0) = x_0 \label{VB_model} \\
C_1&\leq x(t) \leq C_2,\\
P^- &\leq u(t) \leq P^+,\label{VB_constraints}
\end{align}\end{subequations}
where $x(t) \in \mathbb{R}$ denotes the state of charge of the VB at time $t$ with the initial state of charge; $a$ denotes the self-dissipation rate; while the lower and upper energy limits of the VB are denoted by $C_1$ and $C_2$\,, respectively. The regulation signal, $u(t)$ acts as an input to the VB and must always lie within the power limits $P^{-}$ and $P^+$. This simple first-order VB model can be applied to characterize the aggregated flexibility of many building loads and TCLs \cite{hao2015aggregate,hao2017optimal,hughes2016identification}. Note here that, unlike the typical assumption of symmetrical energy limits (i.e. $C_1=-C_2$) in the existing VB identification methods, we allow the lower and upper energy limits to be different. 
%
%
Overall, the vector $\phi = [a, C_1, C_2, x_0, P^- ,P^+]$ denotes the group of VB parameters.

In order to identify the VB parameters, synthetic data are generated concerning the performance of an ensemble of TCLs in responding to a frequency regulation service request. Specifically the synthetic data are generated using simulation models of TCLs described in this section and the regulation signals from PJM \cite{PJM_signals}. This time series evolution of each device with respect to each regulation signal is then used to learn the first order VB system. For the sake of completeness, we also describe briefly the hybrid dynamical models of individual TCLs, used in simulations to generate synthetic data for training and testing of the transfer learning based deep network. Note, however, that the network itself is agnostic of the load models and parameters. 


\subsection{AC Model}
Each AC device is governed by the following temperature dynamics \cite{perfumo2012load,pushpak2018prioritized}. 
\begin{align}
\dot{T}(t) = & - \frac{T(t) - T_a(t)}{C_r R} - \frac{\eta\ p(t)}{C_r},
\label{eq:AC_dynamics}
\end{align}
where 
\begin{align*}
p(t^+) = & \begin{cases} 0, & {\rm if}\,\, T(t) \leq T^{set} - \delta T/2,\\
P, & {\rm if}\,\, T(t) \geq T^{set} + \delta T/2,\\
p(t), & {\rm otherwise}.
\end{cases},
\end{align*}
\begin{tabular}{lcp{6 cm}}
$T(t)$ & - & temperature inside the room \\
$T_a(t)$ & - & outside air temperature \\
$T^{set}(t)$ & - & set point temperature \\
$\delta T$ & - & dead band temperature  \\
$C_r$ &- & thermal capacitance of the room \\
$R$ & - & thermal resistance of the room \\
$\eta$ & - & load efficiency \\
$p(t)$ & - & power drawn by the AC: $0$ when `off'; $P$ when `on'. 
\end{tabular}


\subsection{WH Model}
The below described model is formed by assuming that the water temperature inside the water tank has a uniform temperature, i.e., the water inside the tank is either \textit{all hot} or \textit{all cold} \cite{diao2012electric,pushpak2018prioritized}. 
\begin{align}
\dot{T}_w(t) = & - a(t) T_w(t) + b(t),
\label{eq:WH_dynamics}
\end{align}
where, 
\begin{align*}
a(t) = & \frac{1}{C_w} (\dot{m}(t) C_p + W),\\
b(t) = & \frac{1}{C_w} (p(t) + \dot{m}(t) C_p T_{in} + W T_a(t)),\\
p(t^+) = & \begin{cases} P_w, & {\rm if}\,\, T_w(t) \leq T^{set}_w - \delta T_w/2,\\
0, & {\rm if}\,\, T_w(t) \geq T^{set}_w + \delta T_w/2,\\
p(t), & {\rm otherwise} \end{cases}.
\end{align*}
\begin{tabular}{lcp{6 cm}}
$T_w$ & - & water temperature inside the tank \\
$T_w^{set}$ & - & water temperature set-point \\
$T_{in}$ & - & inlet water temperature \\
$\delta T_w$ & - & dead band temperature \\
$W$ & - & thermal conductance of the tank \\
$C_p$ & - & thermal capacitance \\
$p(t)$ & - & power drawn by the WH: $0$ when `off'; $P_w$ when `on'.  
\end{tabular}
%



\section{Two-stepped training process}
\label{sec7}
Before proceeding into the detailed training process for VB state prediction, we introduce few notations for clarity. $^{1}\mathcal{F}$ and $^{2}\mathcal{F}$ denote the functional representation of Convolution based \textbf{LSTM} network after first and second training process, respectively. $\mathbf{X}$ and $\mathbf{Y}$ denote input and output to the Convolution based \textbf{LSTM} network, during first step of training process. $d$ is used as a historical window for prediction. $^{2}\mathbf{\hat{Y}}$ denotes the prediction of VB state after second step of training process.
\subsection{First step of training process}
This step of the training process involved an unsupervised learning, by utilizing historical VB state and regulation signal of size $d$ (i.e., $d$ number of historical data points consisting VB state and regulation signal at each data point). Given VB state $x$, and regulation signal $u$, input to the unsupervised learning $\mathbf{X}$ is defined as, $\mathbf{X}\triangleq [X_{0},X_{1},\dots,X_{N-1}]$, where $X_{i}\triangleq[x_{i-d}|\dots|x_{i}|u_{i-d}|\dots|u_{i}]$ for all $i=0,1,\dots,N-1$, and output to the unsupervised learning $\mathbf{Y}$ is defined as $\mathbf{Y}\triangleq [x_{1}|\dots|x_{N}]$. The objective of first step of the training process is to learn $^{1}\mathcal{F}$ while minimizing the loss, i.e., $\|\mathbf{Y}-^{1}\mathcal{F}(\mathbf{X})\|_{2}$.
\subsection{Second step of training process}
The purpose of having second step of training process is to mitigate the effect of error accumulation on the forecasting value, over time. For the second step training process, prediction of VB state from previous time step gets used in the next step, along with other historic VB state (we are using $d$ number of historical data points). Upon continuing to forecast in the future, the historic data window keeps getting filled with forecast from previous iterations. This cause a forecasting error accumulation, which results in a divergence of predicted VB state from the actual state magnitude. Algorithm \ref{alg1} which comprises the second step, defines a novel way of mitigating this aforementioned error accumulation.

\begin{algorithm}[!ht]
 \caption{Algorithm of second step training process for virtual battery state prediction}
 \label{alg1}
 \begin{algorithmic}
 \algsetup{linenosize=\small}
 \renewcommand{\algorithmicrequire}{\textbf{Input:}}
 \renewcommand{\algorithmicensure}{\textbf{Output:}}
 \REQUIRE $^{1}\mathcal{F}$, $d$, $\mathbf{X}$, $\mathbf{Y}$
 \ENSURE  $^{2}\mathbf{\hat{Y}}$
  \FOR {i in range($0$,$N$)}
  \STATE count$=-d$
  \IF {$i<d$}
  \FOR {j in range($0$,$2d$)}
  \STATE $\alpha[0,j,0]=\mathbf{X}(j+2d*i)$
  \STATE $\gamma[i,j,0]=\alpha[0,j,0]$
  \ENDFOR
  \STATE $\beta[i]=\mathbf{Y}(i)$
  \ELSE
  \FOR {j in range($1$,$2d$)}
  \STATE $\alpha[0,j,0]=\mathbf{X}(j+2d*i)$
  \STATE $\gamma[i,j,0]=\alpha[0,j,0]$
  \ENDFOR
  \FOR {k in range($0$,$2d$,2)}
  \STATE $\alpha[0,k,0]=\beta[i+count]$
  \STATE $\gamma[i,j,0]=\alpha[0,j,0]$
  \STATE $\text{count}=\text{count}+1$
  \ENDFOR
  \STATE $\beta[i]=^{1}\mathcal{F}(\alpha)$
  \ENDIF
  \ENDFOR\\
  $^{2}\mathbf{\hat{Y}}=\gamma$
 \RETURN $^{2}\mathbf{\hat{Y}}$ 
 \end{algorithmic} 
 \end{algorithm}
 \begin{table*}[!htp]
\centering
\scalebox{0.65}{
\begin{tabular}{|c|c|c|c|c|c|c|c|c|c|c|}
\hline
\multicolumn{1}{|l|}{\multirow{2}{*}{\textbf{Method}}} & \multicolumn{2}{c|}{\textbf{\begin{tabular}[c]{@{}c@{}}Number of\\ devices\end{tabular}}} & \multicolumn{2}{c|}{\textbf{\begin{tabular}[c]{@{}c@{}}Untrained \\ parameters\end{tabular}}} & \multicolumn{2}{c|}{\textbf{\begin{tabular}[c]{@{}c@{}}Pretrained\\ parameters\end{tabular}}} & \multicolumn{2}{c|}{\textbf{Epoch}} & \multicolumn{2}{c|}{\textbf{\begin{tabular}[c]{@{}c@{}}Reconstruction\\ error\end{tabular}}} \\ \cline{2-11} 
\multicolumn{1}{|l|}{} & \textbf{AC} & \textbf{WH} & \textbf{AC} & \textbf{WH} & \textbf{AC} & \textbf{WH} & \textbf{AC} & \textbf{WH} & \textbf{AC} & \textbf{WH} \\ \hline
\begin{tabular}[c]{@{}c@{}}w/o Transfer\\ Learning\end{tabular} & 100 & 0 & 205880 & NA & 0 & NA & 2000 & NA & 0.0028 & NA \\ \hline
\begin{tabular}[c]{@{}c@{}}w/o Transfer\\ Learning\end{tabular} & 112 & 0 & 220280 & NA & 0 & NA & 2000 & NA & 0.0032 & NA \\ \hline
\begin{tabular}[c]{@{}c@{}}with Transfer\\ Learning\end{tabular} & 112 & 0 & 14400 & NA & 205880 & NA & 215 & NA & 0.0028 & NA \\ \hline
\begin{tabular}[c]{@{}c@{}}w/o Transfer\\ Learning\end{tabular} & 0 & 120 & NA & 222980 & NA & 0 & NA & 2000 & NA & 0.0069 \\ \hline
\begin{tabular}[c]{@{}c@{}}w/o Transfer\\ Learning\end{tabular} & 0 & 135 & NA & 241890 & NA & 0 & NA & 2000 & NA & 0.072 \\ \hline
\begin{tabular}[c]{@{}c@{}}with Transfer\\ Learning\end{tabular} & 0 & 135 & NA & 18910 & NA & 222980 & NA & 718 & NA & 0.0061 \\ \hline
\end{tabular}
}
\vspace{20pt}
\caption{Performance comparison between with and without transfer learning for the proposed SAE.}\label{Table1}
\end{table*}

\begin{table*}[!ht]
\centering
\scalebox{0.65}{
\begin{tabular}{|c|c|c|c|c|c|c|c|c|}
\hline
\textbf{Type} & \textbf{\begin{tabular}[c]{@{}c@{}}Number of \\ AC Devices\end{tabular}} & \textbf{\begin{tabular}[c]{@{}c@{}}Number of \\ Water Heaters\end{tabular}} & \textbf{$a$} & \textbf{$C_{1}$} (KWh) & \textbf{$C_{2}$} (KWh) & \textbf{$x_{0}$} & \textbf{$P^{+}$} (KW) & \textbf{$P^{-}$} (KW) \\ \hline
Homogeneous & $100$ & $0$ &2.348& $-140$ & $48$ &  $-128.81$& 423.7656 & -175.0176 \\ \hline
Homogeneous & $112$ & $0$ &2.412  & $-160$ &$50$  &  $-141.65$& 459.6484 & -208.7284 \\ \hline
Homogeneous & $0$ & $120$ & 1.149 & $-420$ & $-36$ & $-397.14$ & 486.0781 & -52.528 \\ \hline
\multicolumn{1}{|l|}{Homogeneous} & $0$ & $135$ & 1.212 & $-451$& $-18$&$-415.73$  & 544.5781  & -60.5353 \\ \hline
\end{tabular}
}
\vspace{20pt}
\caption{Identified VB parameters ($\phi$) for different TCL ensembles}\label{Table2}
\end{table*}

\section{Data and Proposed Method Description}
\label{sec8}
\subsection{Data description}
\label{sec8_1}
We propose VB state $x$ and subsequently VB parameter $\phi$ mentioned in Section \ref{sec6} for an ensemble of homogeneous devices (AC, where each device is satisfying the dynamics in Eq. \eqref{eq:AC_dynamics} and WH, where each device is governed by the dynamics in Eq. \eqref{eq:WH_dynamics}). The regulation signals from PJM \cite{PJM_signals} are considered and scaled appropriately to match the ensemble of ACs and WHs. The devices in each ensemble has to change their state in order to follow a regulation signal. In doing so, while keeping the aggregate power of the ensemble close to the regulation signal, the switching action of the ensemble should not violate the temperature constraints of individual devices. The switching strategy is determined by the solution of an optimization problem (similar to as shown in \cite{hughes2016identification}).

For ensemble of AC devices, a combination of $100$ ACs is considered and the ON and OFF devices at every time instance are identified by solving an optimization problem. This generates the temperature state of each device, $T(t)$, at each time iteration, for $200$ distinct regulation signals. If the ensemble fails to track a regulation signal, then the time-series data is considered up to the point where tracking fails. The outside air temperature and user set-point for each device are assumed to be same in this analysis. 

The power limits of the ensemble are computed through a one-sided binary search algorithm as described in \cite{hughes2015virtual}. The $100$ AC ensemble is simulated for $2$ hours with $1$ sec time resolution, for each regulation signal. For some regulation signals the ensemble violate the power limits $P^-$ and $P^+$ before the $2$ hours running time and the temperature of each AC is only considered, until when the ensemble satisfies the power limit. Finally for making the suitable dataset for applying SAE, we stack temperature of each AC devices, followed by temperature set points for each devices, load efficiency and thermal capacity of each AC devices, by column, and stack the data points for each regulation signals by row. For the selected ensemble, these stacking result in a dataset of dimension $\mathbb{R}^{1440199\times 203}$. To obtain the input stack to SAE for an ensemble of WHs, a similar approach as described above for AC devices is followed. While generating this data, it is assumed that water flow into the WHs is at a \textit{medium} rate.

\subsection{Method description}
The SAE introduced in Section \ref{sec2} is trained on the dataset described in Section \ref{sec8_1} for an ensemble of 100 $AC$ devices. The objective of the training of SAE is to represent the given $203$ dimensional dataset into a $1$ dimensional encoded space, and subsequently transforms the $1$ dimensional encoded representation back into the $203$ dimensional original data space, with tolerable loss. The selected layer dimension of proposed SAE is $203$-$150$-$100$-$50$-$20$-$1$-$20$-$50$-$100$-$150$-$203$, where all the activation functions are linear. Moreover, the variables in $203$ input dimensions are not normalized, to represents the VB state dependency on the input variables. That also motivates the necessity of having unbounded linear activation functions, throughout the proposed SAE.

Moreover, when more AC devices are added to the given $100$ AC devices ensemble, we leverage the proposed \textbf{Net2Net} framework introduced in Section \ref{sec5}, for the retraining and subsequent representation of the VB state for the dataset representing the new ensemble. Obviously the robust way is to retrain the proposed SAE architecture from scratch for the new dataset, but that includes higher computation cost and time. We can utilize the already trained network on $100$ AC devices ensemble, for the new ensemble dataset, which results in significant savings of computation cost and time. 

Finally, we introduced convolution based \textbf{LSTM} network, for forecasting the VB state evolution, given any regulation signal. Given the SAE, only able to represent VB state, for the given time the state of TCL is available, it is required to utilize a deep network for predicting time evolution of VB state, for the ensemble of TCLs. Simultaneously, this proposed convolution based \textbf{LSTM} network can be used to estimate the remaining unknown parameter $a$ in the vector $\phi$, which represents all the parameters related to VB. In the next section, we demonstrate the amalgamation of two proposed deep network framework for an ensemble of $100$, $112$ AC devices, and $120$, $135$ WHs.

\section{Results and Discussion}
\label{sec9}
\begin{figure}
\centering
\includegraphics[width=\columnwidth]
{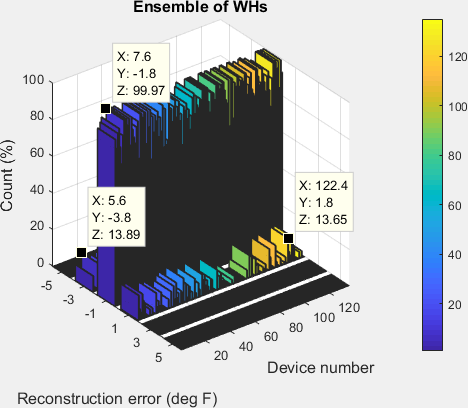} 
\caption{Reconstruction error variation for different WHs in the 135 ensemble, after passing through the proposed SAE. The reconstruction error is within the interval $[-3.8,+1.8]$ deg F.}
\label{fig:sim_plot1}
\end{figure}

\begin{figure}
\centering
\includegraphics[width=\columnwidth]
{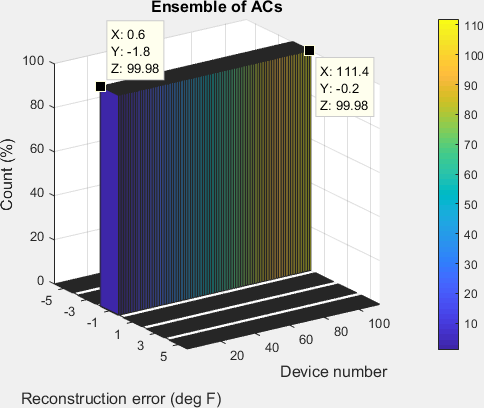} 
\caption{Reconstruction error variation for different ACs in the 112 ensemble, after passing through the proposed SAE. The reconstruction error is within the interval $[-1.8,+1.8]$ deg F.}
\label{fig:sim_plot2}
\end{figure}

We evaluate the performance of our proposed deep network framework on four different ensembles, namely ensemble of $100$ and $112$ AC devices and ensemble of $120$ and $135$ WH devices. Table \ref{Table1} shows the effectiveness of utilizing proposed transfer learning instead of retraining the deep network for every ensemble. Table \ref{Table1} shows an average computation time\footnote{We used Epoch (number of training iterations) from Table \ref{Table1} as a measure of computation time.} savings of $77\%$, when identifying virtual battery states for different ensembles. Figures \ref{fig:sim_plot1} and  \ref{fig:sim_plot2} show the reconstruction error histogram for two different ensembles of WHs and ACs, respectively. Both plots show a very small reconstruction error, which justifies a possibility of utilizing the proposed stacked autoencoder (specifically the trained decoder), to map the virtual battery state to individual device level, for checking the controller performances in device level. Table \ref{Table2} shows calculated virtual battery parameters for four different ensembles, after using the proposed convolution \textbf{LSTM} network. Power limits ($P^{+}$ and $P^{-}$) are calculated using an optimization framework \cite{hughes2015virtual}.

Finally, this proposed framework is shown to be generalizable for different type of TCLs, and also shown to be transferable for different number of devices in any given ensemble. Our future work will involve applying this framework for heterogeneous ensembles.

\bibliographystyle{plain}
\bibliography{sample-bibliography}
\end{document}